\documentclass[10pt]{article} 
\usepackage[preprint]{rlc}

\usepackage{amssymb}            
\usepackage{mathtools}          
\usepackage{mathrsfs}           
\usepackage{graphicx}           
\usepackage{subcaption}         
\usepackage[space]{grffile}     
\usepackage{url}  
\usepackage{algorithm}
\usepackage{algpseudocode}
\usepackage{booktabs}
\usepackage{siunitx}

\let\cite\citep

\title{Simulation-Based Benchmarking of Reinforcement Learning Agents for Personalized Retail Promotions}

\author{Yu Xia  \\
    Yu.Xia@bain.com \\
    Bain \& Company, Inc.
    \And
    Sriram Narayanamoorthy \\
    Sriram.Narayanamoorthy@bain.com\\
    Bain \& Company, Inc.\\
    \And
    Zhengyuan Zhou \\
    zzhou@stern.nyu.edu\\
    New York University\\
    \And
    Joshua Mabry\\
    Joshua.Mabry@bain.com\\
    Bain \& Company, Inc.\\
    }

\begin{document}

\maketitle

\begin{abstract}
The development of open benchmarking platforms could greatly accelerate the adoption of AI agents in retail. This paper presents comprehensive simulations of customer shopping behaviors for the purpose of benchmarking reinforcement learning (RL) agents that optimize coupon targeting. The difficulty of this learning problem is largely driven by the sparsity of customer purchase events. We trained agents using offline batch data comprising summarized customer purchase histories to help mitigate this effect. Our experiments revealed that contextual bandit and deep RL methods that are less prone to over-fitting the sparse reward distributions significantly outperform static policies. This study offers a practical framework for simulating AI agents that optimize the entire retail customer journey. It aims to inspire the further development of simulation tools for retail AI systems.
\end{abstract}
\section{Introduction}
With AI surpassing human-level performance on various benchmarks, such as image classification, basic reading comprehension, and board games, there is an increasing focus on creating autonomous AI agents for specific environments \cite{Ray_Perrault_Jack_Clark_2024}. In retail and e-commerce, advanced reasoning and an understanding of causal relationships are required to make effective product assortment, promotion, and pricing decisions \cite{katsov_introduction_2017}. The combination of these requirements, marketplace dynamics, and the sparsity of customer purchase events across large product catalogs makes it a challenging domain to apply autonomous AI agents. To advance the development of AI in retail, open datasets, and simulation platforms are needed that capture the end-to-end customer experience \cite{Bernardi_Batra_Bruscantini_2021}. Public retail datasets are quite limited, and existing simulation platforms focus on specific problem domains, such as dynamic pricing \cite{serth_interactive_2017} and product recommendations \cite{santana_mars-gym_2020, ie_recsim_2019}. An ideal simulation platform can enable the evaluation of a wide range of marketing agents that optimize customer experiences.

Targeting promotions is already one of the most impactful applications of AI agents; large e-commerce companies, such as Wayfair \cite{Fei_2021}, Booking.com \cite{Kangas_Schwoerer_Bernardi_2021}, Stitch Fix \cite{Glynn_2018}, and Amazon \cite{Kanase2022} have found success using contextual bandits and RL approaches, to decide who gets what offer, when, and over what channel. Enabling this use case has traditionally required large-scale online experimentation programs to collect exploration data and prove the uplift over less sophisticated approaches \cite{Treybig_2022}. Simulation platforms can lower the barrier to adopting RL by enabling the offline development of advanced agents and providing estimates of the potential uplift from deploying them. 

We previously introduced RetailSynth, an interpretable multi-stage retail data synthesizer, and showed that it faithfully captures the complex nature of the retail customer decision-making process over the full journey from choosing to visit a storefront to deciding exactly which product and how much to purchase \cite{xia_retailsynth_2023}. In this work, we extend RetailSynth to enable the training and evaluation of RL agents that target promotions (coupons) to individual customers. We propose an environment where the agent is trained using offline batch data to target store-wide coupons to customers at discrete time steps. In alignment with industry practice, coupons are set up as discrete actions across a range of discount levels and evaluated based on their impact on customer revenue over the evaluation period, while monitoring secondary metrics such as the profit margin impact, the number of categories a customer purchases, and the fraction of customers active at the end of the evaluation period. We characterize the environment using static baseline policies where all customers receive the same coupon and then compare the performance of the baseline policies to personalized policies learned by the RL agents. We segment the customers based on their latent price sensitivity and show that personalized policies typically target less aggressive discounts to less price-sensitive customers. Based on our observation that price-insensitive customers still often receive large discounts, there do appear to be opportunities to improve agent performance on this benchmark. 

To our knowledge, our work is the first to benchmark RL agents on simulated retail customer shopping trajectories. It provides much needed guidance to practitioners on the potential uplift of deploying coupon-targeting agents in a multi-category retail environment. We also provide insights into which customer features effectively summarize the sparse transaction data and a deep dive into metrics to consider prior to deployment. We intend this paper to serve as a blueprint for how to simulate AI agents that optimize the end-to-end retail customer journey. The remainder of our paper is organized as follows: Section \ref{section: environment} gives a detailed overview of the simulation environment; Section \ref{section: experiments} describes the agent training and evaluation experiments; and Section \ref{section: challenges} describes challenges and directions for future work.

\section{Simulation environment}\label{section: environment}
\begin{figure*}[htbp]
    \centering
    \includegraphics[width=1.0\linewidth]{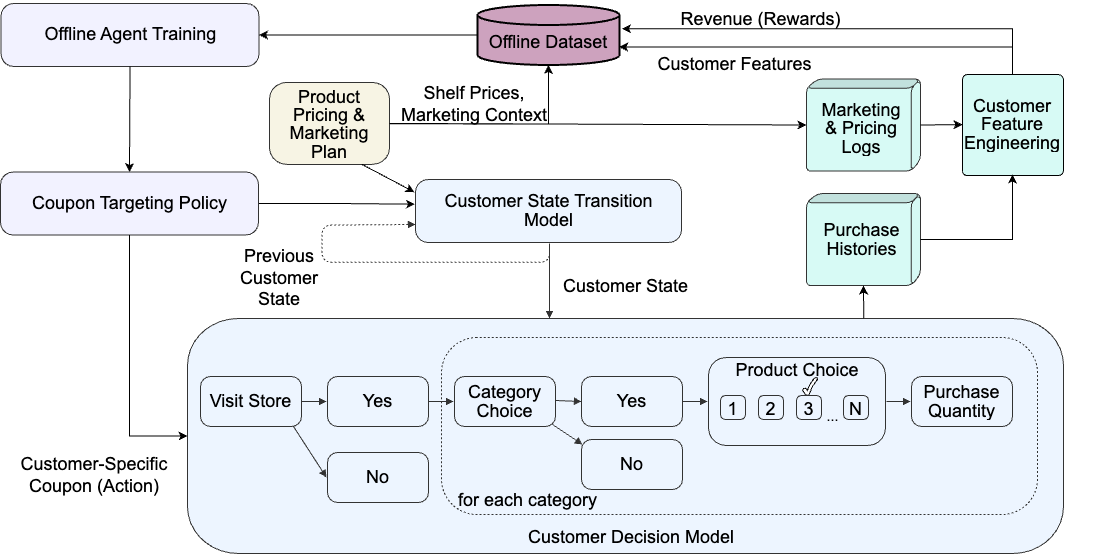}
    \caption{Data flow in RetailSynth environment for evaluating coupon-targeting agents.}
    \label{fig:env-agent-diagram}
\end{figure*}
We leverage the RetailSynth framework to simulate the shopping behavior of a cohort of customers choosing from a large product catalog covering multiple categories. This model was previously calibrated on a public grocery dataset \cite{dunnhumby_complete_2014} and shown to generate realistic, synthetic data (with the KS-statistic < 0.2 for each of the decision stage choice distributions). As shown in Figure \ref{fig:env-agent-diagram}, we integrate the customer decision model with a simulation environment that facilitates training and evaluating offline reinforcement learning agents. The customer decision model is sensitive to changes in the customer state that evolve in response to marketing and pricing decisions. Based on the context of these decisions and the customer purchasing activity, the environment history is summarized at each time step in the simulation to form the batch dataset that is used for agent training. Learned policies are then deployed in the simulation environment, and their performance is measured based on accumulated revenue and other secondary metrics such as category penetration and customer retention.

\subsection{Customer decision model}
The customer decision model is a four-stage decision model that covers the decision to visit the store, to make a purchase within a specific category, which product to buy from a chosen category, and how much of that product to purchase. At each time step in the simulation, the probability of customer $u \in U$ purchasing  quantity $Q$ of product $i \in I$ in category $j \in J$ is given by:
\begin{align*}
&P(Q_{ui}=q,S_u,C_{uj},B_{ui})\\
&=P(S_u)P(C_{uj} | S_u) P(B_{ui}|S_u,C_{uj})P(Q_{ui}=q | S_u,C_{uj},B_{ui}) 
\end{align*}
where $S_u$, $C_{uj}$, $B_{ui}$, and $Q_{ui}$ indicate the binary outcome for each of the listed decision stages. The latent customer state is defined by the product, category, and store visit utilities and is responsive to pricing and marketing decisions. Customer state transitions are captured in the following equations by the inclusion of lagged variables (specifically in the store visit model). The product utility that underlies all decision stages is defined as
\begin{align}\label{eq:product_utility_w_endog}
    \mu^{prod}_{uit} \sim \mathbf{\beta_{ui}^x} \mathbf{X_{uit}} + \beta_{ui}^{z} Z_{i} + \beta_{ui}^w log(P_{uit}) 
\end{align}
$\mathbf{X_{uit}}$ represents observable, time-varying features that are customer- and product-specific (e.g. digital display advertising). $Z_{i}$ is an unobserved factor, such as brand strength, that affects both prices and demand. Note the price sensitivity coefficient $\beta^w_{ui} \sim  \beta_u^w \beta_i^w$ where $\beta_u^w$ and $\beta_i^w$ are customer-specific and product-specific factors.

Each product has a shelf price that is either the base price or a discounted price. The final purchase price $P_{uit}$ is the shelf price, adjusted to account for any personalized coupons the customer chooses to redeem, and is given by $P_{uit} = (1 -D^{coupon}_{uit}*\mathbb{I}^{coupon}_{uit})P^{shelf}_i$. Shelf pricing is assumed to follow a high-low strategy and is simulated using a two-state hidden Markov model as described in \citet{xia_retailsynth_2023}. 

We model a cohort of customers, acquired at the same time period and evolve the probability to revisit the firm based on both outbound marketing and in-store browsing activity. The store visit probability model is an auto-regressive model given by
\begin{align}
P(\mathbb{I}_{ut}) &=
    \begin{cases}
        1 & \text{if } t = 1, \\
        (1-\theta_{u}) \frac{exp(\mu^{store}_{ut})}{ 1 + exp(\mu^{store}_{ut})} + P(\mathbb{I}_{u(t-1)}) \theta_{u} & \text{if } t > 1,
    \end{cases} \label{eq:main_equation}\\
    \intertext{where}
    \mu^{store}_{ut} &= \gamma_0^{store} + \mathbb{I}_{u(t-1)} \gamma_1^{store} SV_{u(t-1)} + \gamma_2^{store} X_{ut}^{store} \label{eq:mu_store}\\
    SV_{ut} &= \log \sum_{j \in J} \exp(CV_{ujt}) \label{eq:store_preference_score}\\
    CV_{ujt} &= \log \sum_{k \in J_j} \exp(\mu^{prod}_{ukt}) \label{eq:category_preference_score}
\end{align}
The store visit utility, $\mu^{store}$ depends on the customer's latent propensity $\gamma_0^{store}$ to revisit the store, outbound marketing activity $X^{store}$, and the effect of browsing activity in the prior time step, $SV$. We make the simplifying assumption here that all products are in the customer's consideration set for each visit but would recommend relaxing this assumption for larger assortments. 

Once the customer has decided to visit the store, category purchase decisions are assumed to be independent Bernoulli trials, where the probability of purchase is given by $\exp(\gamma_{0j}^{cate} + \gamma_{1j}^{cate} CV_{ujt}) / (1 + \exp(\gamma_{0j}^{cate} + \gamma_{1j}^{cate} CV_{ujt}))$. Product choice decisions follow a multinomial logit choice model given by $\exp{(\mu^{prod}_{uit})}/ \sum_{k \in J_j} \exp{(\mu^{prod}_{ukt}})$. For a realized product choice, the purchase quantity is given by a shifted Poisson distribution with the form $\lambda_{uit} ^{q-1}\frac{exp(-\lambda_{uit})}{(q-1)!}$ where $\lambda_{uit} = \exp(\gamma^{prod}_{0i} + \gamma^{prod}_{ui}\ \mu^{prod}_{uit})$.

\subsection{Agents}\label{subsection: agent}
Agents integrated with the RetailSynth environment can optimize product assortment, pricing, or marketing decisions. Here, we focus on personalized marketing agents, leveraging contextual bandit and deep RL algorithms and train those agents off-policy on summarized histories of customer purchases \cite{zhu_pearl_2023} to target customer-specific coupons. 

The offline batch training dataset $\mathcal{D}$  comprises tuples $(H_{u, t-1}, A_{u, t-1}, R_{t})$ and is of length $dim(U) \times T$. Summarized customer purchase histories, $H_t$ are obtained by applying a feature engineering function $f$ to raw observations $(O_{t=0} \ldots O_{t})$. An observation $O_t$ is defined as
\begin{align}
    O_t &= (\mathbf{Q}_{t-1}, \mathbf{P}^{shelf}_t, X^{store}_t, \ \mathbf{X_{t}})
\end{align}
where $\mathbf{Q}_{t-1} = (Q_{i(t-1)}\ for\ i\in I)$ represents the purchase activity in the previous time period, $\mathbf{P}^{shelf}_t = ((1 - D_{it})P_{i}^{base}\ for\ i \in I)$  the product shelf price, $X^{store}_t$ the store marketing features, and $\mathbf{X}_t = (X_{it}\ for\ i\in I)$ the product marketing features.
    
The contextual features used to summarize the customer purchase history, $H_t$, and their relative importance are shown in Appendix \ref{appendix:context_feature_selection}. The action space consists of discrete coupons with discount values, $D^{coupon}$, in the range of $[0, 1)$ that apply to all products $i \in I$. The bandit agents are configured to use the revenue as the reward, $R_{t} = \sum_{i \in I} R_{it} = \sum_{i \in I} P_{it} Q_{it}$ where $P_{it}$ and $Q_{it}$ refer to the product price and quantity sold. Deep RL agents optimize the cumulative revenue over the full trajectory, $V_{t} = \sum_{\tau=1}^t \delta^{t - \tau} R_{\tau}$ where $\delta$ is the discount factor.

\section{Experimental results and discussion}\label{section: experiments}
In our previous work \cite{xia_retailsynth_2023}, we showed that the RetailSynth customer decision model produces highly variable short-term customer purchasing behavior and long-term store loyalty under different pricing scenarios due to the customers having different price sensitivities. We hypothesized that personalized promotions would provide revenue uplift in this environment by targeting discounts to customers based on their latent willingness to pay. In this study, we designed experiments to both validate the environment's suitability for its intended purpose and size the potential revenue uplift from using RL agents to learn personalized promotion policies.

We built our simulation workflows in Python 3.10, with key dependencies on NumPyro \cite{phan2019composable}  for customer choice modeling and TensorFlow Agents \cite{TFAgents} for reinforcement learning. To ensure that we collected sufficient agent training data, we set up a cloud workflow on AWS for parallel computation, leveraging Batch for auto-scaling compute with R4 memory-optimized EC2 instances and S3 for storage \cite{awsbatch, awsec2}. To keep compute costs reasonable, we capped the number of customers in the simulation to 100,000. We also reconfigured simulation parameters to increase the average store visit probability to \~90\% and decrease the size of the product catalog to 2,514. See Appendix \ref{appendix: env params} for more details. We did profile the agent training workflows on G5 GPU compute instances; however, the observed speed increase was <50\%, and we did not find it cost-effective to move our workload to the GPU. The entire workflow for the subsequent experiments consumed approximately 1,950 CPU-hours in total.

\subsection{Benchmark policies}
\begin{figure}[htbp]
  \begin{minipage}{0.5\textwidth}
    \includegraphics[width=\textwidth]{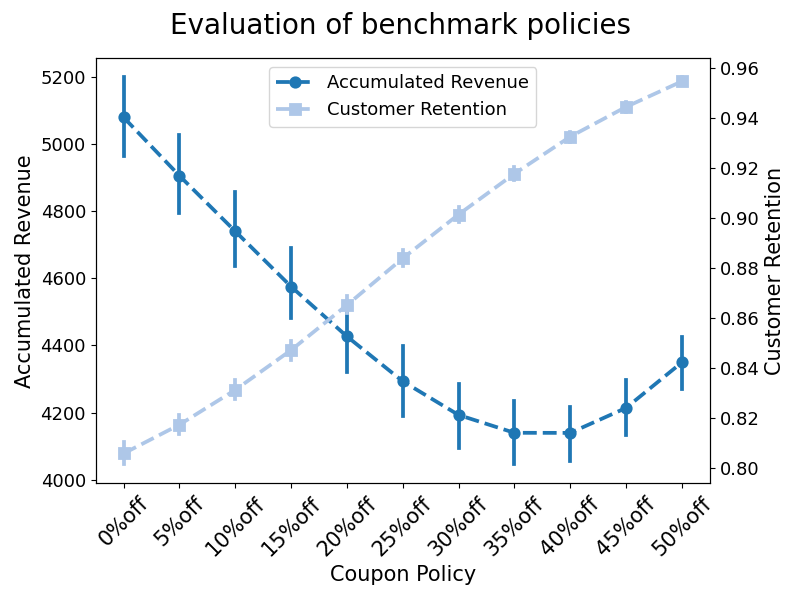}
  \end{minipage}\hfill
  \begin{minipage}{0.4\textwidth}
    \caption{Accumulated revenue and customer retention when applying fixed coupon policies to 100 separate simulations of 100 customers over 70 time steps. Metrics collected from last 20 time steps to match evaluation period used for agent training.} \label{fig:scenario_exploration}
  \end{minipage}
\end{figure}

To verify that the simulation environment presented non-trivial trade-offs between short-term revenue and long-term customer loyalty, we first conducted simulations of static benchmark policies. In these simulations, we gave all customers the same coupon at each time step and then measured the accumulated revenue and customer retention rate after 70 time steps (Figure \ref{fig:scenario_exploration}). We did observe a trade-off, with revenue and retention rates showing opposing trends for lower coupon discounts from 0\% to 40\%. The customer retention rate continued to increase for coupon discounts up to 50\%, while the trend in accumulated revenue reversed. Based on these observations, we expected that an optimal policy would assign a range of coupon values to specific customers to increase customer retention and maximize overall revenue. For benchmarking learned policies, we used the static policy of 0\% coupon value since it yielded the highest overall revenue.

\subsection{Agent training and evaluation}
We first collected an offline training dataset, leveraging a random collection policy, where the probability of choosing different coupon levels for each customer is uniformly distributed (Algorithm \ref{alg:offline_data}). The offline dataset comprised data from 100,000 customers over $T=50$ time steps, equivalent to roughly one year of purchasing history. In our distributed computing environment, we optimized memory usage by setting the batch size $B=100$ customers, requiring $N_{batch}=1000$ parallel simulations. To train the agents, we followed Algorithm \ref{alg:train_eval_one_agent}, training the agent on the offline dataset covering $t= 1, \dots, T$ and then evaluated their performance by resuming the simulation at time step $t = T +1$ and collecting $T_{eval}=20$ additional observations. 

Our overall goal was to determine which agents could give the maximum performance with optimal hyper-parameters. For hyper-parameter tuning, we sampled $N_{tune}=20$ different hyper-parameter configurations using the Tree-structured Parzen Estimator in Optuna \cite{akiba_optuna_2019} and then selected the best configuration based on the average accumulated revenue from $N_{agent}=3$ independent training and evaluation runs. For final benchmarking with optimal hyper-parameters (Table \ref{tab:hyper-params}), we followed the same agent training procedure, while increasing  $N_{agent}$ to $10$ and setting $N_{eval}$ to $10$. See Appendix \ref{appendix:hypeparameter_tuning_log} for more details. 
\begin{table}[htbp]
\centering
\fontsize{8pt}{8pt}\selectfont
\begin{tabular}{
  l
  l
  S[table-format=1.1]
  S[table-format=1.1]
  S[table-format=1.1]
  S[table-format=1.1]
  S[table-format=1.1]
}
\toprule
\textbf{Agent Type} & \textbf{Policy} & \textbf{Objective} & \multicolumn{4}{c}{\textbf{Secondary metrics}} \\
\cmidrule(lr){4-7}
& & \shortstack{Accumulated\\Revenue} & \shortstack{Accumulated\\Demand} & \shortstack{Customer\\Retention} & \shortstack{Category\\Penetration} & \shortstack{Coupon\\Discount} \\
\midrule
 Benchmark & 0\%off \textcolor{red}{*} & 1.12 & 0.63 & 0.90 & 0.63 & $/$ \\
 & & \pm0.06 & \pm0.02 & \pm0.01 & \pm0.02 &  \\
\midrule
Contextual Bandit & LinTS \textcolor{red}{*} & \textbf{1.26} & 1.39 & 0.99 & 1.39 & 0.69 \\
& &  $\mathbf{\pm0.06}$ & \pm0.04 & \pm0.01 & \pm0.04 & \pm0.02 \\
\cmidrule(lr){2-7}
 & LinUCB \textcolor{red}{*} & \textbf{1.26} & 1.39 & 0.99 & 1.39 & 0.69 \\
 & &  $\mathbf{\pm0.06}$ & \pm0.04 & \pm0.01 & \pm0.04 & \pm0.02 \\
\cmidrule(lr){2-7}
 & NB \textcolor{red}{*} & 1.07 & 0.88 & 0.97 & 0.88 & 0.64 \\
 & & \pm0.05 & \pm0.03 & \pm0.01 & \pm0.03 & \pm0.03 \\
\midrule
Reinforcement Learning & PPO \textcolor{red}{*} & $\mathbf{1.26}$ & 1.31 & 0.98 & 1.31 & $\mathbf{0.58}$ \\
 & & \pm0.06 & \pm0.04 & \pm0.01 & \pm0.04 & $\mathbf{\pm0.02}$ \\
\cmidrule(lr){2-7}
& DQN & 0.96 & \textbf{1.50} & $\mathbf{1.05}$ & $\mathbf{1.50}$ & 1.71 \\
 & & \pm0.04 & $\mathbf{\pm0.05}$ & $\mathbf{\pm0.01}$ & $\mathbf{\pm0.05}$ & \pm0.02 \\
\bottomrule
\end{tabular}
\caption{Summary of agent performance metrics normalized relative to a random policy (with a value of 1 indicating performance equal to the random policy). The mean and standard error of each metric are reported based on 10 learned policies, evaluated over 10,000 customers and across 20 time steps. A red asterisk denotes that the agent performs significantly better than the random agent, as determined by a t-test on the accumulated revenue distributions. Bold font indicates the best performing agents for each metric.
\label{tab:agent_performance_summary}}
\end{table}

We trained and evaluated a wide range of agents: linear contextual bandits (Linear Thompson Sampling (LinTS), Linear Upper Confidence Bound (LinUCB)); a neural contextual bandit (Neural Boltzmann (NB)); and deep reinforcement learning methods (Proximal Policy Optimization (PPO) and Deep Q-Network (DQN)). The performance metrics we computed included accumulated revenue, accumulated demand, customer retention, category penetration, and redeemed coupon discount value (Table \ref{tab:agent_performance_summary}). Reflecting common industry practice, we report accumulated revenue as the primary objective and consider the other objectives secondary. In the case where multiple agents yield similar revenues, we would recommend selecting an agent for deployment that performs best along the secondary metrics most relevant to the firm. For example, if the firm is trying to grow, then increasing customer retention and category penetration might be considered more important than minimizing the coupon discount rate. 

Our results indicate that all the agents, except DQN, effectively learn to target coupons more effectively than a random policy. In this scenario, it is important not only to beat a random policy but also to outperform static benchmarks. Comparing against the best baseline policy of offering everyone a 0\% coupon discount, only LinTS, LinUCB, and PPO agents show improved revenue performance. Looking at the secondary metrics, we find that the PPO agent generated the highest level of revenue, while minimizing the average coupon discounts. In a real-world setting, we would recommend deploying this policy based on the observed performance. 

The relative performance of the different agents reflects important characteristics of the environment and training workflow that we have built here. First, the strong performance of the linear contextual bandit agents implies a simple linear structure in terms of rewards and state-action relationships. In addition, the strong performance of the PPO agent and poor performance of the DQN agent may be explained by two factors. First, the random collection policy is a relatively strong policy, and so a policy gradient method like PPO can reliably learn an effective policy. Second, the reward distribution is sparse (due to the customers purchasing only a small number of items in the catalog) and difficult to model accurately without overfitting, which may explain the poor performance of the NB and DQN agents that leverage neural reward models. 

To provide additional guidance to practitioners, we performed a sensitivity study where we decreased the size of the training dataset and evaluated the performance of the top 3 agents, LinUCB, LinTS, and PPO. As shown in Appendix \ref{appendix: sensitivity analysis}, we observed only about a 1\% decrease in revenue performance for the contextual bandit agents (LinTS and LinUCB) when decreasing the number of customers in the training set by 10x, while the PPO agent performance decreased about 7\%. It is noteworthy that all of the agents provide revenue uplift, even with just 10,000 active customers---a fact that might surprise practitioners accustomed to case studies involving larger firms with millions of customers. It also reinforces the importance of using simpler algorithms over more complex ones early in the algorithm development process, where achieving lift over existing baselines as quickly as possible is the objective versus at a later point in time where the focus may shift to maximizing the performance of a proven decisioning system.

\subsection{Analysis of learned policies}
\begin{figure}[htbp]
  \centering
  \includegraphics[width=0.95\textwidth]{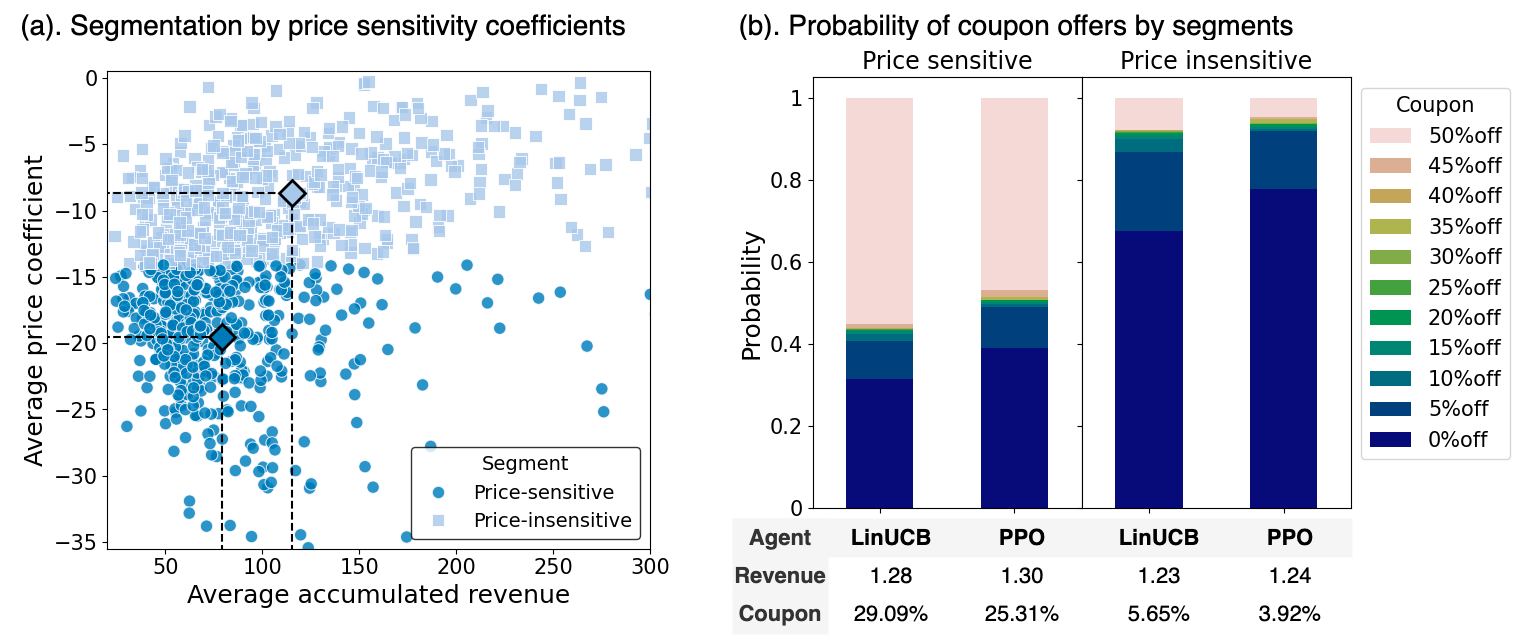}
  \caption{(a) The average accumulated revenue of customer segments with different price sensitivities. Average price coefficient given by $\sum_i \beta_{ui}^w / dim(I)$. Customers in the price-sensitive and price-insensitive segments have average price coefficients greater than or less than the median. Diamond indicators show the mean values of the price coefficient and revenue for each segment. (b) Coupon offer probability distributions by segment shown in the bar chart. The table shows average accumulated revenue of each agent normalized relative to the random policy and the actual average coupon discount value.}
  \label{fig:seg-panel}
\end{figure}
We conduct an in-depth analysis of the driving factors of agent performance and estimate how much potential improvement more advanced algorithms might achieve on this benchmark. Leveraging the interpretable nature of the simulations, we divide customers into two equal-sized segments based on their latent price sensitivity (Figure \ref{fig:seg-panel}a). From a business point-of-view, the price-insensitive segment is more valuable because the customers, on average, generate nearly 50\% more revenue, and so there is a real cost to offering them larger coupons than required to maximize long-term revenue. 

We chose to carefully analyze the policies of LinUCB and PPO agents based on their strong performance and differences in algorithmic complexity. We observe in Figure \ref{fig:seg-panel}b that both agents target the coupons as expected the majority of the time. The agents primarily allocated 50\% off coupons to price-sensitive customers and 0\% off to the price-insensitive segment. However, seemingly non-optimal coupons are also frequently offered to customers in both segments. 

In Appendix \ref{appendix:mean_reward_dist_by_segment}, we show the mean reward distribution of each offer for each segment. There is a clear separation between the best and worst actions for the price-insensitive segment, yet the bandit agent, which should choose actions with the highest reward, still chooses the 50\% coupon discount almost 10\% of the time for this segment. We observe a similar action allocation distribution for the PPO agent, which implies that the shortcomings of the policies are more a result of the input data rather than formulation of the optimization objective or model architecture. Improved performance then is likely to come from increasing the volume of training data and the richness of the customer features used to train the agents.

\section{Challenges and Future Directions} \label{section: challenges}
Adoption of AI agents in traditional industries like retail will be greatly accelerated by developing open datasets, benchmarking suites, and simulation platforms that enable reproducible evaluation and iterative development of new methods. In this work, we demonstrate how to leverage RetailSynth simulations of customer shopping behavior to benchmark the performance of coupon-targeting agents. This approach can scale to large customer bases and product assortments and be extended for a variety of important use cases from marketing message optimization to assortment selection. We also recognize that there are a number of important challenges to overcome:
\begin{enumerate}
    \item \textbf{Retail environment}: The real-world retail environment includes omnichannel customer experiences, marketplace effects, and inventory constraints. These are important factors to add to the environment to make it more realistic. In addition, learning systems may already be in place for high-value use cases. Future work should also include having learned policies as the collection policy for agent training.  
    \item \textbf{Customer decision modeling:} RetailSynth is based on a simple mechanistic model of customer behavior that we designed to ensure a heterogeneous customer response to pricing decisions and to be highly interpretable and tunable to different business scenarios. However, transferring agent learnings from simulation to real-world deployments requires a more accurate modeling approach. 
    \item \textbf{Feature engineering}: We worked with a modest feature space to minimize the computational overhead of training and evaluating multiple agents. RL frameworks should be extended to include more efficient and well-integrated pre-processing methods and minimize the need for hand-building features. 
    \item \textbf{Action space}: Within the chosen use case of targeting coupons, additional complexity can be added to more closely mimic real-world scenarios. For example, we can build agents that select from many offers specific to individual categories or product bundles. For other use cases, such as dynamically altering the product assortment, the action spaces are combinatorial and will present scalability challenges.
    \item \textbf{Agent tuning}: Obtaining reasonable performance with RL methods requires extensive domain understanding and hyper-parameter tuning due to the sensitivity of the algorithms to the choice of hyper-parameters. The development of more robust algorithms and more automated training routines will allow for faster experimentation and greater adoption.
\end{enumerate}

Addressing these challenges will help drive reinforcement learning forward and make it more practical to deploy AI agents in traditional industries such as retail. Our ultimate goal is to enable retailers to offer more deeply personalized and satisfying shopping experiences. 
\bibliography{main}
\bibliographystyle{rlc}
\appendix
\section{Simulation environment parameters}\label{appendix: env params}
The simulator was calibrated previously to match the Complete Journey dataset\cite{xia_retailsynth_2023}. The store visit probability was $\sim 50\%$, and the conditional product purchase probability was $\sim 3\%$, leading to a very sparse dataset. To increase the purchase frequency in these simulations, compared to the previous work, we modified Scenario V from the prior work to have an empirical discount depth of $\sim 30\%$. We also decreased the store visit model parameters to $\gamma_0^{store} \sim Gumbel(-2, 0.1)$ and $\gamma_2^{store}\sim Uniform(0.004, 0.006)$ to retain a broad range of retention rates across customers. In addition, we reduced the size of the product catalogs from 26,176 products to 2,514.

\section{Simulation workflow}\label{appendix: sim protocol}
We describe here the simulation workflow from the collection of training data to agent training and hyper-parameter optimization and agent evaluation. We first describe the components and parameters of the simulation workflow. They key outputs of the simulator are training and evaluation datasets that are composed of customer trajectories. A trajectory for customer $u$ is denoted as $\mathcal{D}_T = \{ (H_{u, t-1}, A_{u, t-1}, R_{t}) \}_{t=1}^{T}$. Dropping the subscript $u$ for convenience, the key variables in our simulations are as follows:  

\begin{itemize}
    \item $\mathbb{O}$: the observation space. $O_t \in \mathbb{O}$ denotes the observations the environment generates at time $t$.
    \item $\mathbb{S}$: the state space. $S_t \in \mathbb{S}$  denotes the hidden state of the environment at time $t$
   \item $\mathcal{A}$: the available action space. $A_t \in \mathcal{A}$ denotes the action the agent chooses at time $t$.
    \item $R$: the reward function, representing the immediate impact of an action, $R_t = R(S_{t-1}, A_{t-1})$. We assume the reward for action at time $t-1$ will be provided to the agent at time $t$. 
    \item $\Omega$: the environment. $\Omega(S_t)$ denotes the environment status at time $t$
\end{itemize}

In Table \ref{tab:simulation_parameters}, we show the full set of parameters that define our agent training and evaluation workflows. Algorithm \ref{alg:offline_data} describes the protocol to collect offline data for agent training. Algorithm \ref{alg:train_eval_one_agent} describes how that offline data is used to train and evaluate multiple agents, such that we can obtain bootstrap estimates of the statistics describing agent performance. Algorithm \ref{alg:hyperparam_tuning} describes how we we used Optuna \cite{akiba_optuna_2019} to optimize the hyperparameters for agent training. 

\begin{table}[H]
\centering
\fontsize{9pt}{6pt}\selectfont
    \caption{Simulation Parameters}
    \label{tab:simulation_parameters}
    \begin{tabular}{ll}
    \toprule
    Parameter & Description\\
    
    \midrule
    
    \multicolumn{2}{l}{\textbf{Algorithm \ref{alg:offline_data}}} \\
        $\theta_{env}$ & Parameters to initialize an environment \\
        \addlinespace
        $B$ & Number of customers in one batch\\
        \addlinespace
        $N_{batch}$ & Number of customer batches\\
        \addlinespace
        $T$ & Length of one episode\\
        \addlinespace

    \midrule
    
    \multicolumn{2}{l}{\textbf{Algorithm \ref{alg:train_eval_one_agent}}} \\

        $\rho_{agent}$ & Parameters to initialize an agent \\
        $N_{train}$ & Number of training epochs \\
        $N_{eval}$ & Number of evaluation episodes \\
        $T_{eval}$ & Length of one evaluation episode \\

    \midrule

    \multicolumn{2}{l}{\textbf{Algorithm \ref{alg:hyperparam_tuning}}} \\

        $\rho_{low}$ & Lower bound of parameter range while tuning \\
        $\rho_{high}$ & Upper bound of parameter range while tuning \\
        $N_{tune}$ & Number of tuning trials \\
        $f$ & Function to compute optimization objective

    \end{tabular}
\end{table}

\begin{algorithm}[H]
\caption{Pseudocode for offline data collection}\label{alg:offline_data}
\begin{algorithmic}[1]
\State Input $\theta_{env}, B, N_{batch}, T$
\State Initialize $\mathcal{D} \gets \{ \}$
\For {$n = 1, \cdots, N_{batch}$}
    \State Initialize environment $\Omega$ with $\theta_{env}$ for $B$ customers
    \State Initialize trajectory $D \gets \{ \}$
     \State Initialize random policy $\pi$ s.t. $\pi(a) > 0$, $\forall a \in \mathcal{A}$ 
     \State Generate an episode $\{O_0, A_0, R_1, 
     \cdots, O_{T -1 }, A_{T -1 }, R_{T}\}$
     \State Summarize observations
     \For {$t=0, \cdots, T-1 $}
        \State $H_t \gets f(O_0,\cdots, O_t)$
        \State $D \gets D \cup (H_t, A_t, R_{t+1})$
        \EndFor
\State $\mathcal{D} \gets \mathcal{D} \cup \{(D, \Omega)\}$
\EndFor
\end{algorithmic}
\end{algorithm}

\begin{algorithm}[H]
\caption{Pseudocode for agent training and evaluation}\label{alg:train_eval_one_agent}
\begin{algorithmic}[1]
\State Input $\rho_{agent}, N_{train}, \mathcal{D}$
\State Input  $N_{eval}, T_{eval}, N_{agent}$
\For {$N_{agent}$}
\State Initialize policy $\pi$ with $\rho_{agent}$ s.t. $\pi(a) > 0$, $\forall a \in \mathcal{A}$
\For {$n = 1, \cdots, N_{train}$}
    \State Sample $D, \Omega$ from $\mathcal{D}$
     \State Update $\pi$ w.r.t $D$
\EndFor
\State Initialize $\mathcal{D}_{eval} \gets \{\}$
\For {$n = 1, \cdots, N_{eval}$}
    \State Generate an episode using $\Omega$ and $\pi$
    \State $D_{eval} \gets \{H_{T + 1}, A_{T + 1}, R_{T + 2},\cdots, H_{T + T_{eval} - 1}, A_{T + T_{eval} - 1}, R_{T + T_{eval}}\}$
     \State $\mathcal{D}_{eval} \gets \mathcal{D}_{eval} \cup \{D_{eval}\}$
\EndFor
\EndFor
\end{algorithmic}
\end{algorithm}

\begin{algorithm}[H]
\caption{Pseudocode for agent hyperparameter tuning}\label{alg:hyperparam_tuning}
\begin{algorithmic}[3]
\State Input $\rho_{low}, \rho_{high}, N_{tune}, f$
\State Input $N_{train}, \mathcal{D}, \theta_{env}, B, N_{eval}, T_{eval}$
\State Initialize Tree-structured Parzen Estimator (TPE) sampler $\varrho$
\State Initialize optimal evaluation trajectory container $\mathbb{V}_{opt} \gets \{\}$
\For {$n = 1, \cdots, N_{tune}$}
    \State Sample $\rho$ from $\varrho$
    \State $\mathcal{D}_{\rho} \gets$ Apply Algorithm\ref{alg:train_eval_one_agent}
    with $\rho$ and $N_{train}, \mathcal{D}, \theta_{env}, B, N_{eval}, T_{eval}$
    \State $V_{\rho} \gets \mathbb{V}_{opt} \cup \{f (\mathcal{D}_{\rho})\}$ 
\EndFor
\State $\rho_{opt} \gets \text{argmax}_{\rho} \mathbb{V}_{opt}$
\end{algorithmic}
\end{algorithm}

\section{Selection of context features}\label{appendix:context_feature_selection}
\begin{figure}[htbp]
    \centering
    \includegraphics[width=0.7\textwidth]{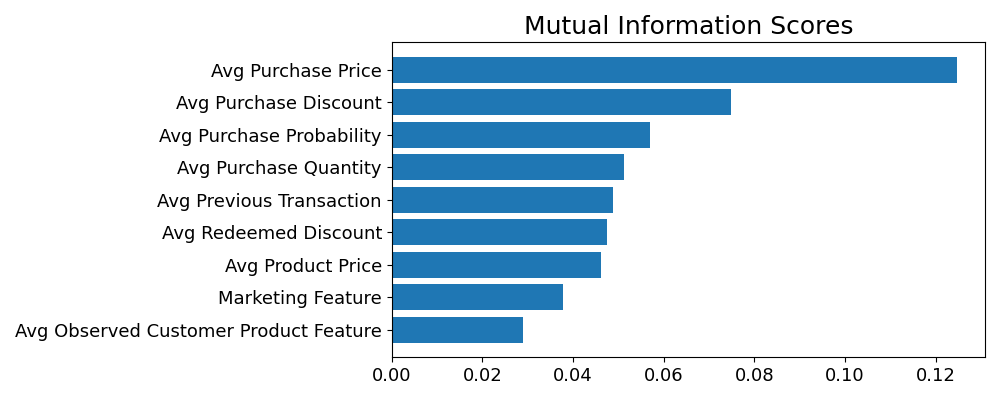}
    \caption{Mutual Information scores of manually prepared features ranked in descending order.}
    \label{fig:feature-importance}
\end{figure}

In our study, we follow the typical industry practice of engineering features to enhance the predictive power of models and aid interpretability. We prepared aggregated features from the transactional data and marketing activity logs, such as average purchase price, average purchase discount, etc. To verify the relevance of these features, we computed using mutual information scores \cite{Kraskov_2004} with transaction reward as the continuous target variable (Figure \ref{fig:feature-importance}) and verified the information content was non-negligible. Based on this analysis, we included all the engineered features for agent training.

\section{Hyperparameter tuning} \label{appendix:hypeparameter_tuning_log}
\begin{table}[htbp]
\centering
\fontsize{8pt}{3pt}\selectfont
\begin{tabular}{lllll}
\toprule
\textbf{\shortstack{Agent\\ Type}} & \textbf{Policy} & \textbf{Hyper-parameter} & \textbf{Search Range} & \textbf{\shortstack{Optimal\\ Value}} \\
\midrule
Contextual & LinTS & alpha & $\in [0.1, 0.9]$ & 0.7387\\
\addlinespace
Bandit & & gamma & $\in [0.1, 0.9]$ & 0.8119\\
\addlinespace
\cmidrule(lr){2-5}
 & LinUCB & alpha & $\in [0.1, 0.9]$ & 0.8483\\
\addlinespace
 & & gamma & $\in [0.1, 0.9]$ & 0.1088\\
\addlinespace
\cmidrule(lr){2-5}
 & NB & optimizer.learning\_rate & $\in [0.001, 0.05]$ & 0.0250\\
\addlinespace
 & & temperature & $\in [0.1, 0.9]$ & 0.1665\\
\addlinespace
 & & hidden\_layers\textcolor{red}{*}  & $(8, 2), (16, 4), (32, 8)$ & (8, 2)\\
\addlinespace
\midrule
Reinforcement & PPO & optimizer.learning\_rate & $\in [0.001, 0.01]$ & 0.0064\\
\addlinespace
Learning & & importance\_ratio\_clipping & $\in [0.1, 0.9]$ & 0.7849\\
\addlinespace
 & & discount\_factor & $\in [0.1, 0.9]$ & 0.8199\\
\addlinespace
 & & entropy\_regularization & $\in [0.0, 0.5]$ & 0.1013\\
\addlinespace
 & & num\_epochs & $5, 6, \cdots, 15$ & 12\\
\addlinespace
 & & actor\_net.fc\_layer\_params & $(8, 2), (16, 4), (32, 8)$ & (32, 8)\\
\addlinespace
 & & value\_net.fc\_layer\_params & $(8, 2), (16, 4), (32, 8)$ & (32, 8)\\ 
\addlinespace
\cmidrule(lr){2-5}
 & DQN & optimizer.learning\_rate & $\in [0.001, 0.01]$ & 0.0050\\
\addlinespace
 & & gamma & $\in [0.1, 0.9]$ & 0.8792\\
\addlinespace
& & epsilon\_greedy & $\in [0.1, 0.9]$ & 0.1552\\
\addlinespace
 & & q\_network.dense\_layer.units\textcolor{red}{*} & $16, 32, 64$ & 16\\
\addlinespace
 & & q\_network.lstm\_layer.units\textcolor{red}{*} & $4, 8$ & 4\\

\bottomrule
\end{tabular}
\caption{Setup of hyper-parameter tuning log for each policy using Optuna. Parameters without red asterisk are all native parameters required by tf-agents to initialize corresponding agents. Refer to the paper source code for detailed definitions of highlighted hyperparameters otherwise. Initial parameter values are suggested by uniform distribution if the search range is an interval or categorical distribution if the search range lists out the potential candidates. The optimal value is extracted from the configuration leading to the maximum average accumulated revenue
\label{tab:hyper-params}}
\end{table}
We ran hyper-parameter tuning jobs using Optuna \cite{akiba_optuna_2019} with the optimization objective of maximizing average accumulated revenue. For each policy, we sampled $N_{tune}=20$ parameter configurations and gather objective value using Algorithm \ref{alg:train_eval_one_agent} with $N_{eval}=10, T_{eval}=20, N_{agent}=3$. Within each run, we sampled with replacement from the offline dataset, keeping the training dataset size unchanged at 100,000 customers. Due to memory limitations, we trained LinTS, LinUCB using two mini-batches with 50, 000 customers in each. We trained Neural Boltzmann, PPO, DQN agents with mini-batches of 2,000 customers, 1,000 training epochs, and an early-stopping callback with the tolerance of 15 time steps to monitor the loss convergence.

\section{Agent training sensitivity analysis} \label{appendix: sensitivity analysis}
\begin{figure}[htbp]
    \centering
    \includegraphics[width=0.7\textwidth]{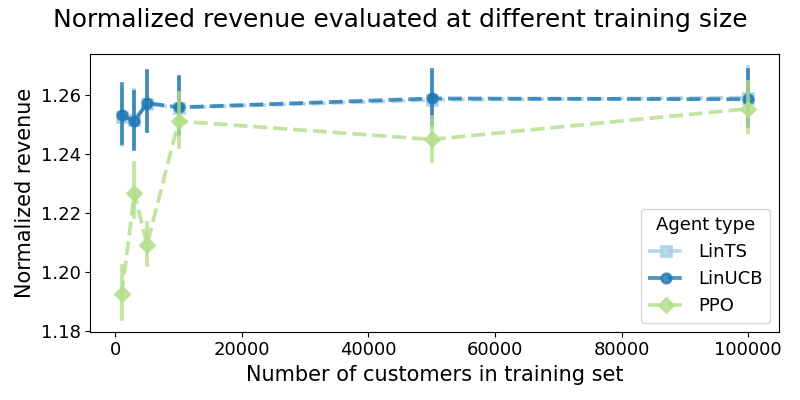}
    \caption{Impact of training data size on agents' accumulated revenue normalized by the random agent.}
    \label{fig:training-size}
\end{figure}

We further investigated the impact of varying training data sizes on agent performance by training and testing agents with offline data trajectories with 1000, 5000, 10000, 50000, and 100000 customers. In Figure \ref{fig:training-size}, we observed that the size of the training data had a modest impact on performance for linear agents, LinTS and LinUCB, with slight fluctuations in the mean of accumulated revenue across different training sizes. Conversely, the performance of the PPO agent exhibited a substantial increase with larger training sizes. The PPO agent trained with more extensive datasets consistently outperformed the one trained with smaller datasets, indicating a positive correlation between training data size and agent effectiveness. This finding suggests that while linear bandits can efficiently explore optimal arms with relatively small datasets, more complex agents like PPO benefit significantly from larger training datasets, enabling them to learn more intricate decision-making policies and improve overall performance.

\section{Offer revenue by segment}\label{appendix:mean_reward_dist_by_segment}

\begin{figure}[H]
    \centering
    \includegraphics[width=0.6\textwidth]{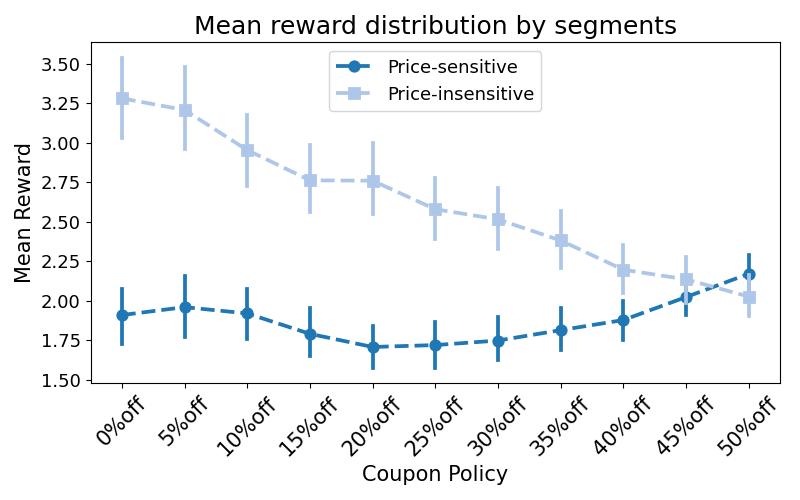}
    \caption{Distribution of mean reward by customer segments and coupon offers in the offline dataset collected under Algorithm \ref{alg:offline_data}}
    \label{fig:mean_reward_distribution}
\end{figure}
\end{document}